\documentclass[conference]{IEEEtran}
\usepackage{amsmath,amssymb,amsfonts}
\usepackage{graphicx}
\usepackage{hyperref}
\usepackage{array}
\usepackage{multirow}
\usepackage{fancyhdr}
\newcolumntype{x}[1]{>{\centering\arraybackslash\hspace{0pt}}p{#1}}
\begin{document}

\title{Improving Machine Learning Based Sepsis Diagnosis Using Heart Rate Variability}

\author{\IEEEauthorblockN{Sai Balaji}\IEEEauthorblockA{\textit{The Woodlands College Park High School}\\The Woodlands, TX, United States\\\texttt{sai.s.balaji@gmail.com}}
\and
\IEEEauthorblockN{Christopher Sun}\IEEEauthorblockA{\textit{Stanford University} \\Stanford, CA, United States \\\texttt{chrisun@stanford.edu}}
\and
\IEEEauthorblockN{Anaiy Somalwar}\IEEEauthorblockA{\textit{University of California, Berkeley} \\ Berkeley, CA, United States \\\texttt{anaiy@berkeley.edu}}
}

\maketitle

%right notice
\thispagestyle{plain}
\fancypagestyle{plain}{
\fancyhf{} % clear all header and footer fields
\fancyfoot[L]{979-8-3503-5663-2/24/\$31.00~\copyright2024~IEEE} % change copyright notice here if required
\renewcommand{\headrulewidth}{0pt}
\renewcommand{\footrulewidth}{0pt}
}

\begin{abstract}
The early and accurate diagnosis of sepsis is critical for enhancing patient outcomes. This study aims to use heart rate variability (HRV) features to develop an effective predictive model for sepsis detection. Critical HRV features are identified through feature engineering methods, including statistical bootstrapping and the Boruta algorithm, after which XGBoost and Random Forest classifiers are trained with differential hyperparameter settings. In addition, ensemble models are constructed to pool the prediction probabilities of high-recall and high-precision classifiers and improve model performance. Finally, a neural network model is trained on the HRV features, achieving an F1 score of 0.805, a precision of 0.851, and a recall of 0.763. The best-performing machine learning model is compared to this neural network through an interpretability analysis, where Local Interpretable Model-agnostic Explanations are implemented to determine decision-making criterion based on numerical ranges and thresholds for specific features. This study not only highlights the efficacy of HRV in automated sepsis diagnosis but also increases the transparency of black box outputs, maximizing clinical applicability.
\end{abstract}


\section{Introduction}
Sepsis poses a significant global health issue with a high mortality rate and economic burden \cite{seymour2020derivation} \cite{iwashyna2016sepsis}. Sepsis is characterized by a dysregulated host response to infection, causing fever and an abnormal white blood cell count \cite{rhodes2020surviving}. Accurate diagnosis of sepsis is critical for reducing mortality rates and substantial healthcare costs \cite{rudd2020global} \cite{kleinpell2013impending}. Though advances like the quick Sequential Organ Failure Assessment (qSOFA) may enhance early sepsis detection, the frequent vital sign monitoring and reassessment required adds significant burdens to nursing workload, in return for already-questionable timeliness and accuracy \cite{islam2023machine}. Heart rate variability (HRV), which reflects autonomic nervous system activity and is extractable from heart-monitoring devices like ECG, emerges as a promising non-invasive tool for early sepsis detection due to its ability to detect dysregulation before clinical symptoms appear \cite{shashikumar2020early,ahmad2018impending,de2018heart,leon2021early}. 

Bedoya et al. (2020) and van Wijk et al. (2023) employ machine learning and continuous ECG monitoring for sepsis prediction, harnessing both invasive and non-invasive patient data to advance early sepsis detection standards \cite{bedoya2020machine} \cite{vanwijk2023predicting}. In contrast, our research zeroes in on exploiting only HRV metrics for \textit{non-invasive} sepsis diagnosis, offering a streamlined, patient-centric approach. 

Sendak et al. (2020) successfully implement ``Sepsis Watch,'' a deep learning based sepsis detection tool requiring both invasive and non-invasive clinical data. Integration of this tool into everyday clinical practice demonstrates the practicality of enhancing sepsis management with advanced machine learning techniques \cite{sendak2020real}. Henry et al. (2015) develop TREWScore, a predictive model for septic shock utilizing a wide array of physiological and laboratory data, outperforming traditional methods in early detection \cite{henry2015targeted}. Building upon these two works, our study aims to demonstrate that similar state-of-the-art models can operate on HRV metrics exclusively, resulting in a simpler feature space and a more explainable and effective method for early sepsis detection in a clinical environment. 

% Wee et al. (2020) provided a comprehensive narrative review concentrating on the application of heart rate and HRV in diagnosing and managing pediatric sepsis, highlighting the limitations of conventional heart rate measurements and advocating for HRV analysis as a valuable, non-invasive enhancement for sepsis identification and prognosis in pediatric patients \cite{wee2020narrative}. Aligning with Wee et al.'s insights, our research advances the application of HRV analysis for sepsis detection in a broader demographic, underlining its potential for universal clinical utility. 

Additionally, in comparison to van Wijk et al.'s analysis of 171 emergency organ dysfunction cases, our study is the first to examine a more extensive dataset of 4,314 patient cases \cite{vanwijk2023predicting}. Our novel approach spearheads statistical analyses, feature selection, and model construction strategies to uncover relevant relationships between 57 HRV-derived metrics and sepsis. We also present case studies demonstrating enriched understanding of model decision-making that may be crucial for physician-mediated environments. 
% significantly enriching the scope of non-invasive sepsis monitoring across diverse clinical environments with LIME (Local Interpretable Model-agnostic Explanations) based explanations to enhance interpretability and understanding of the predictive model's decisions. 

\section{Methods}
\begin{figure}[hbt!]
    \centering
    \includegraphics[width=0.43\textwidth]{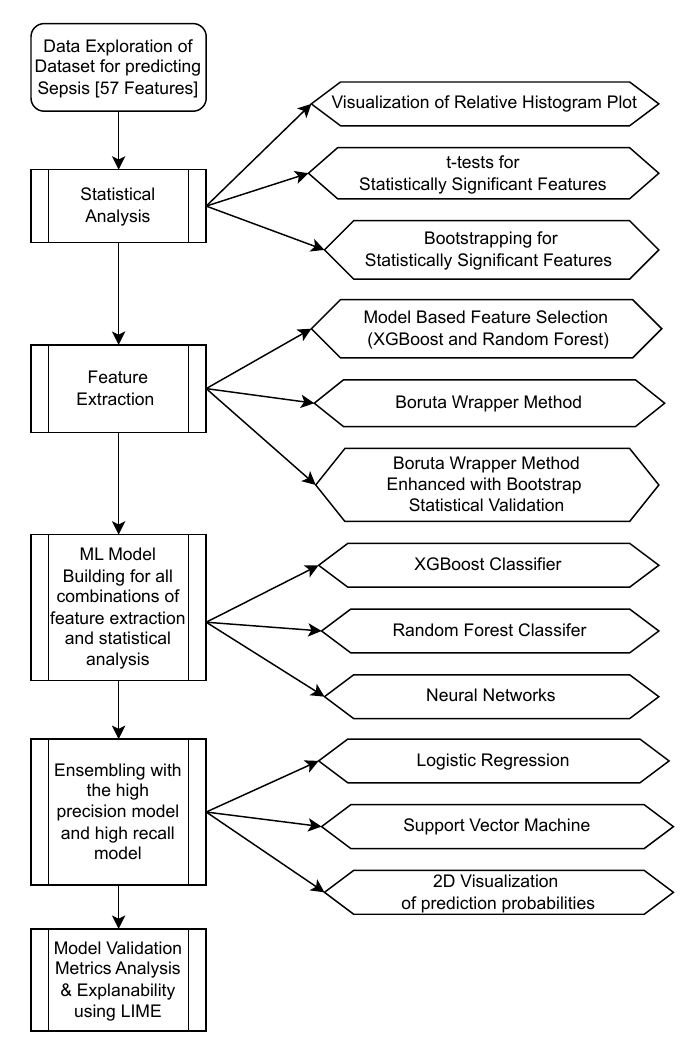}
    \caption{Experimental Design}
    \label{flow}
\end{figure}

Fig. \ref{flow} encapsulates this study's methodology, portraying experimental design, data acquisition, statistical analysis, feature selection, model construction, ensembling mechanisms, and model explainability under the umbrella of sepsis diagnosis.

\subsection{Data Availability Statement}
The dataset used in this research is titled ``Derivation and Validation of Heart Rate Variability-Based Scoring Models for Continuous Risk Stratification in Patients with Suspected Sepsis in the Emergency Department,'' authored by Kuan-Fu Chen from Chang Gung University College of Medicine \cite{chen2020hrv}. The dataset comprises 3776 non-Sepsis-3 records and 538 Sepsis-3 records, where \textit{Sepsis-3} is the latest consensus definition on sepsis and septic shock. The dataset does not include any personally identifiable information, and ethical considerations such as patient inclusion criteria are managed by the original authors. Because of class imbalance, for both feature selection and model construction, we split data into training (80\%) and testing (20\%) sets using stratified sampling with a random state to ensure balanced class distribution. We also implement class weights as loss function coefficients to combat class imbalance. Feature selection, model construction, ensembling techniques and hyperparameter optimization are conducted exclusively on the training set to prevent data leakage and detect overfitting, with model performance validated on the independent test set. The 57 HRV features with descriptions identified for sepsis diagnosis are detailed in Table \ref{Feature_sel}. 

To our knowledge, no previous studies have exclusively focused on HRV data/metrics for the early detection of sepsis using machine learning techniques. However, this means that direct comparisons and benchmarking against existing methods is not feasible, underscoring the need for dataset standardization allowing for comprehensive evaluations.

\subsection{Statistical Analysis}
\subsubsection{Relative Histogram Plot and Preliminary Assessment}
We preliminarily assess feature relevance by visually comparing the distribution of HRV features between individuals with and without sepsis. We then conduct \textit{t}-tests and nonparametric bootstrapping for each HRV feature to identify statistically significant differences in mean values between patient subgroups. The nonparametric bootstrap enhances the reliability of the \textit{t}-test by addressing the normality assumption. These statistical analyses establish a foundation for identifying HRV features most relevant to sepsis, setting the stage for more complex machine learning model development.

\subsection{Feature Selection}
We test three separate feature selection mechanisms and discuss a final method that incorporates elements from all.
\subsubsection{Model-based Feature Selection}
Model-based feature selection utilizes the intrinsic ``importance scores'' from ensemble tree-based methods, such as random forest and gradient boosting, to identify and retain HRV features that significantly contribute to predicting sepsis. This approach captures influential predictors and excludes irrelevant noise from the dataset. The random forest classifier used for feature selection employs 1,000 trees with a maximum depth of 10, while the XGBoost classifier is configured with a learning rate of 0.01, 1,000 estimators, and a maximum depth of 4.

\subsubsection{Boruta Wrapper Method}
The Boruta algorithm, a wrapper method, systematically identifies relevant HRV features by generating shadow features, comparing their importance with actual features' importance through a random forest classifier, and only selecting features that consistently show stronger signals than the highest-ranking shadow feature \cite{kursa2010boruta}. We employ the Boruta method with 100 trials, each using a maximum depth of 20 for the random forest classifier.

\subsubsection{Statistical Selection}
Statistical selection uses bootstrapping to identify statistical significance, determining features based on sheer numerical relevance for sepsis diagnosis. 

As our ultimate method, we apply Boruta in conjunction with statistical validation, first using the Boruta algorithm to identify potentially relevant features, then using statistical selection to filter for features surpassing a significance threshold. Our hybrid method yields an optimized feature set with high predictive ability, thereby improving the accuracy and dependability of the sepsis predictive model.

\subsection{Model Construction}
The model construction phase encompasses data preprocessing to standardize feature scales followed by the training of Extreme Gradient Boosting (XGBoost), Random Forest, and Neural Network models. For XGBoost and Random Forest, we optimize decision thresholds by selecting the $\beta$ parameter (of the F-beta score) that maximizes F1 score. Moreover, we handle imbalanced classes within the dataset by implementing class weights as loss function coefficients. The following are design specifications for our models: 
\begin{itemize}
    \item \textbf{Random Forest Classifier}: Employs 1,000 trees with a maximum depth of 10, and class weight set to ``balanced.''
    \item \textbf{XGBoost Classifier:} Employs maximum of 1,000 estimators, learning rate of 0.01, maximum depth of 4, and gamma of 0.9.
    \item \textbf{Neural Network:} Contains four dense layers populated with 64 hidden units each. Each layer uses Glorot Normal initialization, batch normalization, and ReLU activation, followed by a 0.4 dropout rate to lessen overfitting. Trained with Adam optimization using a learning rate of 0.005, compiled with binary cross-entropy loss and metrics including precision, recall, and F1 score.
\end{itemize}

\subsection{Ensembling Techniques to Couple Model Decision-Making}
After the development of individual models, we implement an ensemble method to leverage the strengths of individual predictive models for diagnosing sepsis, aiming to enhance overall predictive performance by combining models with high precision and recall. Prediction probabilities from models excelling in one of precision and recall are stored in a data frame, forming the training data for subsequent ensemble strategies.

\subsubsection{ML Ensembling}
This ensemble approach integrates the output probabilities of a high precision model (HPM) and a high recall model (HRM) as input features, then trains a machine learning model on these probabilities with the objective of enhancing overall F1 score. Standardization of the input features is performed using standard z-score normalization. We apply Logistic Regression with default hyperparameters as a proof of concept and a Support Vector Classifier configured with a radial basis function kernel, C=1.0, degree=4, gamma=``scale,'' and class weight set to ``balanced'' to the task.

\subsubsection{Manual Ensembling}
The manual ensembling method introduces a custom, rule-based strategy to combine predictions from models with high precision and recall, delivering a final sepsis diagnosis based on conditions of agreement or disagreement between the models. By assigning compound prediction statuses through a custom function, this technique examines model consensus and leverages the comparative advantages of high precision and high recall models. 

\subsection{Model Explainability}
In an effort to bridge the gap between complex machine learning models, simple diagnostic outputs, and practical clinical applicability, this study applies Local Interpretable Model-agnostic Explanations (LIME) to illuminate the decision-making processes of the highest-performing sepsis prediction model \cite{ribeiro2016should}. LIME provides granular insights into individual predictions, highlighting the relevance of specific HRV features in the model's conclusions, thereby offering clues into the influence of features on the model's decisions. LIME provides local explanations, focusing on a small region of the input space around specific predictions, and may not capture the global behavior of the model. Therefore, its interpretations should be considered as a supplementary tool rather than a definitive representation of how complex models operate for clinicians and healthcare practitioners. For this study, LIME explanations are generated for 20\% of the test data to better interpret model predictions.  
% This approach not only enhances trust in the predictive model but also fosters a collaborative environment in the adoption of AI technologies in healthcare, emphasizing the importance of interpretability and transparency in machine learning applications for real-world medical diagnostics.

\section{Results}
\subsection{Statistical Analysis}
\begin{figure}[hbt!]
    \centering
    \includegraphics[width=0.43\textwidth]{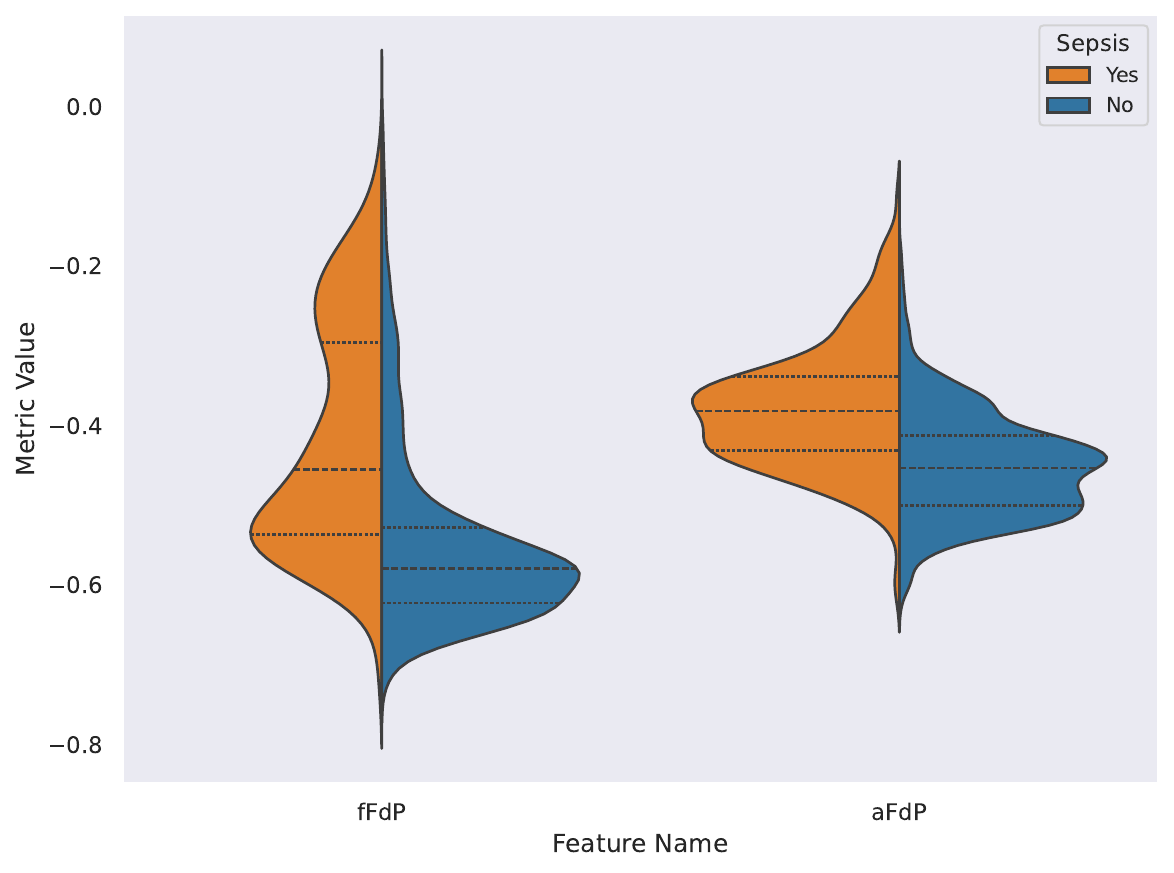}
    \caption{Feature Distributions by Sepsis Status}
    \label{stats}
\end{figure}

\begin{figure}[hbt!]
    \centering
    \includegraphics[width=0.43\textwidth]{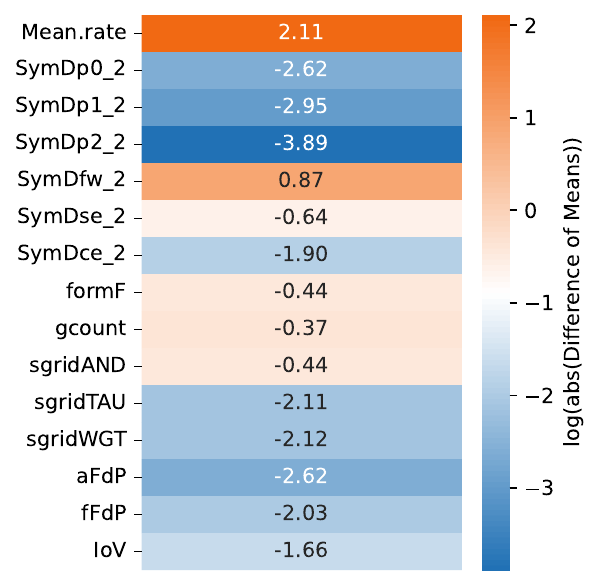}
    \caption{Top 15 Statistically Significant Sepsis Features Based on Bootstrapping Difference Of Means}
    \label{diff-of means}
\end{figure}

Fig. \ref{stats} shows the distributions of two selected features for healthy and infected patients. We can see that the difference in mean values is visually noticeable; such observations guide our statistical feature selection framework. We further quantify the degree of phenotype difference using the $t$-statistic and $p$-value for each HRV feature. Statistical analysis reveals that mean heart rate, though elementary, is a strong predictor for sepsis. Alongside mean heart rate are features like Shannon entropy and ``forbidden words,'' derived from a symbolic dynamics representation of time-series heart data. Fig. \ref{diff-of means} displays the logarithm of the absolute value of the difference in means for the fifteen most statistically significant features in the dataset.

\subsection{Feature Selection and Importance}
Utilizing the XGBoost and Random Forest model-based feature selection methods result in the identification of 52 and 25 key features, respectively, with a significant overlap, underscoring crucial HRV metrics detailed in Table \ref{Feature_sel}. The Boruta algorithm identifies 42 features, and a subsequent statistical-based bootstrapping analysis pinpoints 41 of these 42 features as statistically significant. These 41 features are collectively noted as Boruta-bootstrap features as in Table \ref{Feature_sel}. Our multi-faceted feature selection process not only underscores the complexity of sepsis pathophysiology but also highlights the diverse HRV metrics essential for accurate prediction models.

\subsection{Model Performance Metrics}
As a result of exhaustive testing of hyperparameters, feature sets, and machine learning architectures for the sepsis diagnosis task, we create three tables summarizing the performance metrics for all architectures tested. Tables \ref{F1}, \ref{Precision} and \ref{Recall} capture the nuances of the relationship between said settings and model performance. Namely, XGBoost is the better overall performer compared to Random Forest, and the Boruta-bootstrap feature selection mechanism tends to result in the best model when optimizing for precision and recall -- individually and in tandem. On the other hand, using random forest's intrinsic feature importance scores seem to lead to weak performance, which may not necessarily mean that these importance scores are irrelevant, but may signify the added value of fine-tuning. Overall, we are unable to configure a model yielding both a precision and recall higher than 0.8; the best ML model is an XGBoost with bootstrap-determined features, achieving a F1 score of 0.745.

\begin{figure}[hbt!]
    \centering
\includegraphics[width=0.43\textwidth]{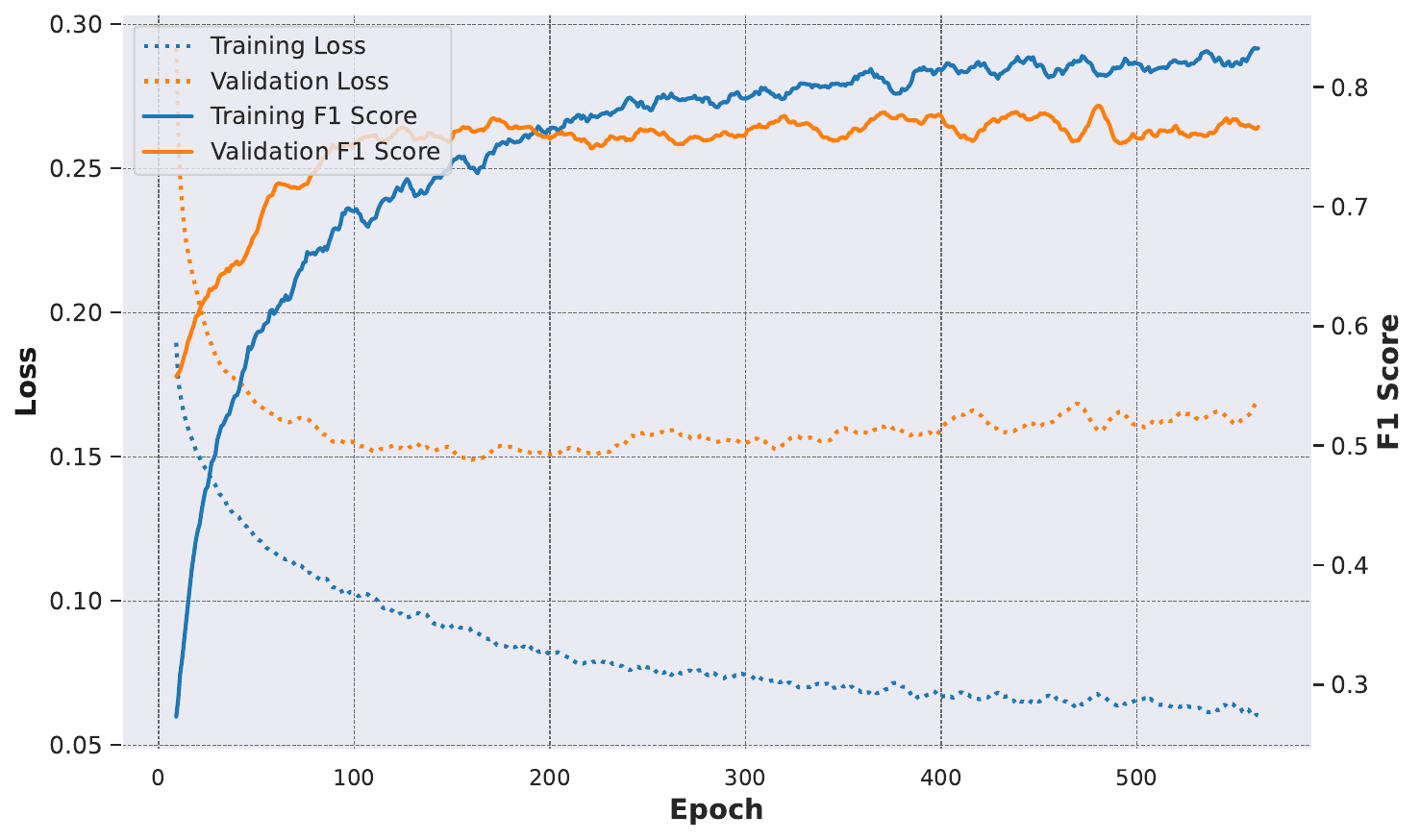}
    \caption{Neural Network Learning Curves: Loss and F1 Score}
    \label{Training_NN}
\end{figure}

Notably, the Neural Network outperforms all machine learning models, emerging as a promising HRV-based classifier of sepsis. Utilizing Boruta-bootstrap features, our model achieves a F1-score of 0.805, with a precision of 0.851 and recall of 0.763. We display learning curves in Fig. \ref{Training_NN}, which illustrate convergence after roughly 500 training epochs and mitigation of overfitting. Table \ref{confusion-matrix} displays a confusion matrix for the neural network model's correct and incorrect predictions on the testing data, and Table \ref{res} displays performance metrics.

\begin{table}[hbt!]
\caption{Confusion Matrix from Neural Network Model}
\begin{tabular}{l|l|l|l}
\cline{2-3} \\[-1em]
& Predicted: No Sepsis & Predicted: Sepsis & \\ \hline \\[-1em]
\multicolumn{1}{|l|}{Actual: No Sepsis} & TN = \textbf{926} & FP = \textbf{18} & \multicolumn{1}{l|}{944}     \\ \hline \\[-1em]
\multicolumn{1}{|l|}{Actual: Sepsis} & FN = \textbf{32} & TP = \textbf{103} & \multicolumn{1}{l|}{135} \\ \hline \\[-1em]
& 958 & 121 & \multicolumn{1}{l|}{1079} \\ \cline{2-4} 
\end{tabular}
\label{confusion-matrix}
\end{table}

\subsection{Ensembling Techniques}
In theory, ensembling classifiers together for coupled model decision-making should improve performance \cite{hasan2020diabetes}. Our results weakly support this claim, as we find that the best ensemble does not perform as well as the best fine-tuned individual ML model, but outperforms an off-the-shelf XGBoost baseline. The logistic regression ensemble, merging a high-precision XGBoost model with a high-recall Random Forest model, achieves an F1-score of 0.724. Interestingly, the support vector machine ensemble demonstrates better performance by 0.4\% but has a 11.2\% higher recall and 10.2\% lower precision. Overall, both ensembling approaches yield favorable outcomes summarized in Table \ref{res}. We can think of the input data frame of probabilities as an information-dense, specially-selected set of features; this allows for greater coupling accuracy even if the ensemble is a simple model like logistic regression.

\begin{figure}[hbt!]
    \centering
    \includegraphics[width=0.43\textwidth]{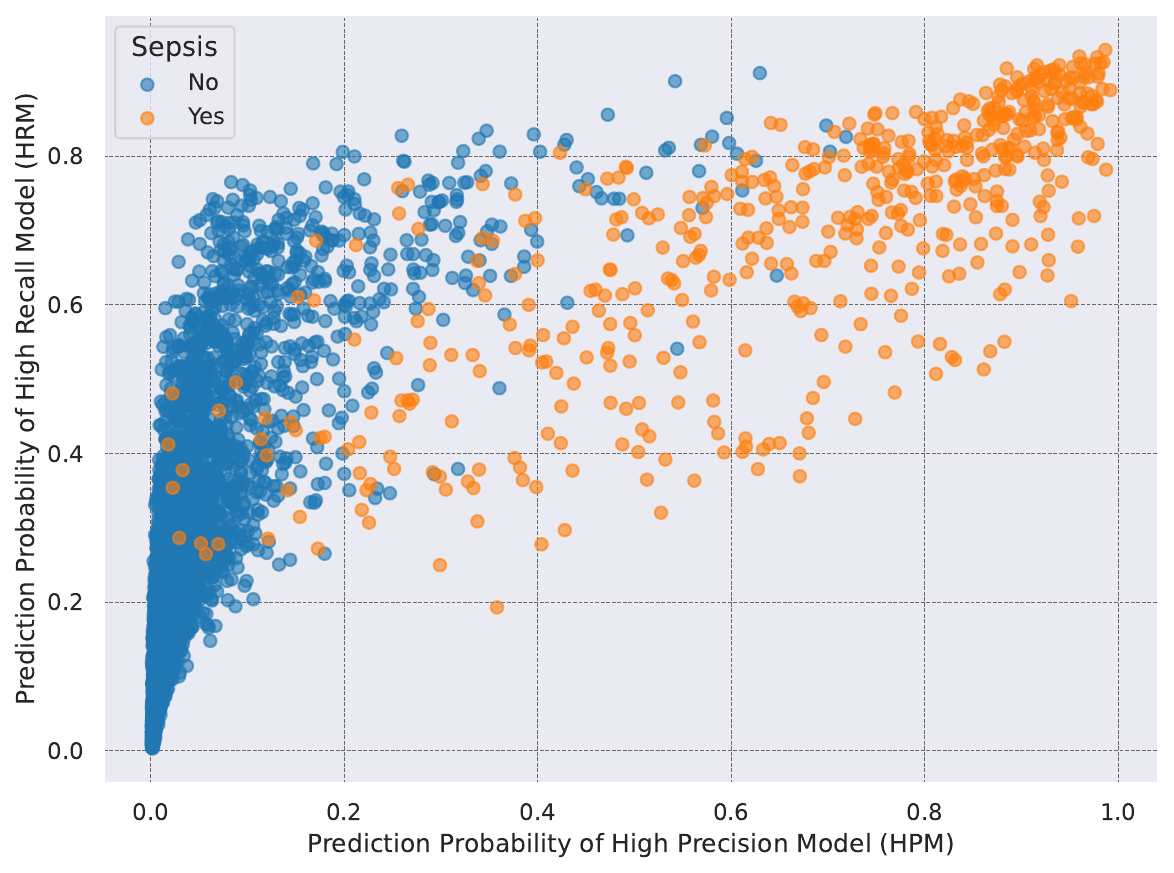}
    \caption{2-D Feature Space: HPM and HRM Prediction Probabilities}
    \label{precisionVSrecall}
\end{figure}

We further visualize the results of ensembling by plotting two sets of probabilities against each other: Fig. \ref{precisionVSrecall} contains prediction probabilities of the high precision model (HPM) on the $x$-axis and prediction probabilities of the high recall model (HRM) on the $y$-axis. Ideally, this plot yields two ``regions'' separable by a straight line (or any other decision boundary that can be learned by an ML model). In our case, there seems to be a faint curve of separation between blue and orange points, but the presence of several outliers prevents the perfect separation of healthy and affected patients. 

The manual ensemble method also reveals a notable recall of 82.5\% alongside a model accuracy of 72.07\%. Regarding clinical implementation, we encounter a 26.76\% classification disagreement rate between the high precision and high recall models. In practice, these patients should be referred for physician diagnosis.

\begin{table}[hbt!]
\caption{Predictive Performance Across Machine Learning Models}
\begin{center}
\begin{tabular}{|l|c|c|c|}
\hline
\textbf{Model}&\textbf{F1 Score}&\textbf{Precision}&\textbf{Recall}\\\hline
Baseline XGBoost & 0.682 & 0.689 & 0.676\\\hline
Ensemble: Logistic Regression & 0.724 & 0.791 & 0.667\\\hline
Ensemble: SVM & 0.728 & 0.689 & 0.778\\\hline
Best ML Model (XGBoost) & 0.745 & 0.760 & 0.731\\\hline
\textbf{Neural Network} & \textbf{0.805} & \textbf{0.851} & \textbf{0.763}\\\hline
\end{tabular}
\label{res}
\end{center}
\end{table}
\subsection{Explainability Insights}
Local Interpretable Model-agnostic Explanations (LIME) for the XGBoost model and Neural Network provide insightful perspectives on how HRV features are differentially weighed in a model prediction. From this, we may be able to untangle the complexity of HRV features and even infer physiological implications.

\begin{figure}[hbt!]
    \centering
    \includegraphics[width=0.43\textwidth]{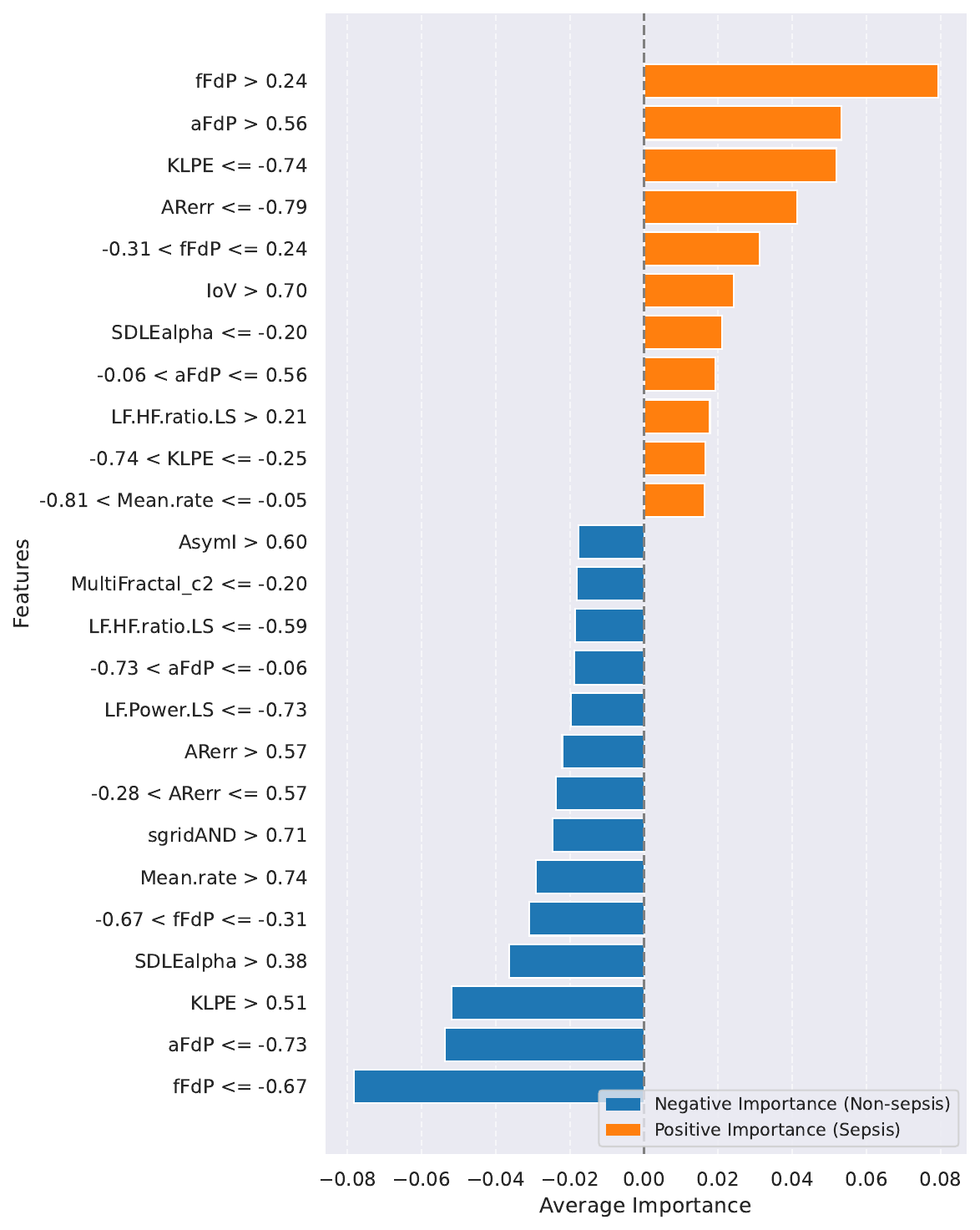}
    \caption{LIME Average Feature Importance on Test Dataset using XGBoost Model}
    \label{explain-xgboost plot}
\end{figure}

\subsubsection{Key Findings from XGBoost (Fig. \ref{explain-xgboost plot})}
\begin{itemize}
\item \textbf{Frequency Domain Measures}: The feature fFdP surfaces with significant importance across various ranges: when $\text{fFdP}>0.24$, the XGBoost model interprets this as an elevated sepsis risk by almost 8\% on average. When $-0.31<\text{fFdP}\leq-0.24$, the severity of increased risk drops to around 3\%. When $-0.67<\text{fFdP}\leq-0.31$, the model reverses its decision-making, adding around 3\% to the healthy diagnosis. Finally, when $\text{fFdP}\leq-0.67$, the model is confident in a decreased sepsis risk by almost 8\%. This case study highlights the nuanced impact of HRV frequency domain measures on the model's sepsis diagnosis. As one can see, the XGBoost model ``discretizes'' the fFdP feature, transforming a continuous feature scale into relevant percentages toward or against a sepsis diagnosis.
\item \textbf{Amplitude and Non-linear Characteristics}: Features such as aFdP and KLPE appear in multiple ranges, indicating the critical role of amplitude variations and non-linear dynamics in HRV analysis, potentially reflecting the autonomic nervous system's regulation.
\item \textbf{Variability Indicators}: IoV (index of variability) and mean heart rate are identified as vital for understanding HRV patterns, as they may signify physiological adaptability and resilience.
\end{itemize}

\begin{figure}[hbt!]
    \centering
    \includegraphics[width=0.43\textwidth]{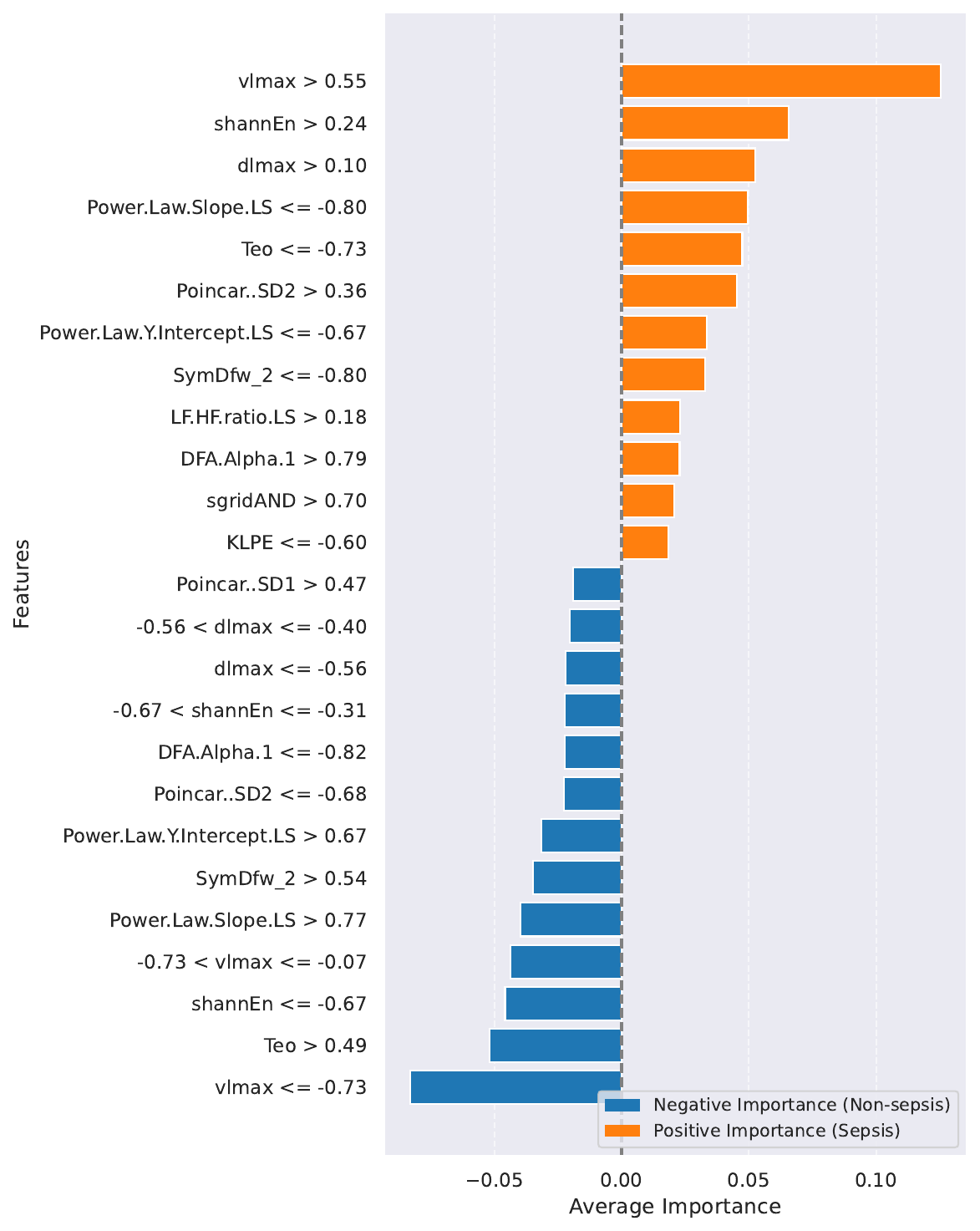}
    \caption{LIME Average Feature Importance on Test Dataset using Neural Network}
    \label{Explain_Neural_Network}
\end{figure}

\subsubsection{Neural Network Model Insights (Fig. \ref{Explain_Neural_Network})}
\begin{itemize}
    \item \textbf{Highlighted Features}: Unlike the XGBoost model, the neural network does not prioritize fFdP, and rather seems to choose a whole new set of features altogether to give prominence to in the decision-making process. For example, the features vlmax, dlmax, shannEn, and Poincar.SD2 have a pronounced association with HRV outcomes.
\end{itemize}

\subsubsection{Comparison}
Analysis from both models, while showcasing some overlap in feature significance (e.g. frequency domain measures, non-linear dynamics), also reveals distinct differences in the importance range of specific features. This discrepancy may arise from inherent differences in how each model approaches the task, with the neural network modeling non-linear relationships for example. 

\section{Discussion}
\subsection*{Advancements in HRV-Based Early Sepsis Detection}
Our comprehensive framework not only highlights the potential for artificial intelligence to significantly advance sepsis diagnostics but also emphasizes the promise of non-invasive monitoring methodologies through heart rate variability. Our initial trials with machine learning models, including Extreme Gradient Boosting (XGB) and Random Forest (RF), with different hyperparameters and feature selection mechanisms reveals a maximum F1 score of 0.745. Introducing a multi-layer neural network within this framework significantly increases this metric, as our neural network achieves a 0.805 F1 score. This suggests nonlinearity between HRV features and sepsis. 

We try to bring clinical utility to this work in two ways. We first offer interpretable explanations that allow physicians and users to understand a model's decision-making criteria. We also prescribe a manual ensembling mechanism where two models, each specializing in one of precision and recall, are locked in a system of checks and balances to provide a final sepsis diagnosis. This manual ensemble is a feasible clinical tool, as instances of model agreement can be trusted (0.825 recall), while instances of model disagreement can simply call for human intervention. This represents a pivotal progression toward integrating machine learning in clinical settings for predictive diagnostics; overall, our study corroborates the claims of van Wijk et al. (2023) regarding the indispensable role of HRV analysis in healthcare outcomes \cite{vanwijk2023predicting}.

\subsection{XGBoost LIME Explanations}
\textbf{Frequency Domain Measures:} Van Wijk et al. find that the average of the NN-interval, ultra-low frequency (ULF), very low frequency (VLF), low frequency (LF), and total power differs significantly between groups with varying degrees of organ dysfunction in sepsis patients \cite{vanwijk2023predicting}. In our case, LIME explanations for the XGBoost model identify frequency domain-related features such as LF.Power.LS and LF.HF.ratio.LS as having considerable importance. This alignment suggests that changes in the frequency domain measures of HRV are indeed indicative of physiological alterations, potentially related to organ dysfunction or sepsis.

% \textbf{Nonlinear Features:} Although \cite{vanwijk2023predicting} did not explicitly mention nonlinear HRV features in their key findings, the significance of non-linear HRV metrics (e.g., KLPE) identified by the LIME explanation for the XGBoost model might offer an additional layer of insight into the complex physiological processes involved in sepsis and its progression. Non-linear analyses can capture the complexity and irregularity of heart rate dynamics that may not be fully encompassed by traditional HRV measures.

\subsection{Neural Network LIME Explanations}
\textbf{Continuous Monitoring Capability:} The utility of continuous ECG monitoring highlighted by van Wijk et al. aligns with the importance of dynamic features such as Poincar.SD2 and shannEn identified by the Neural Network's LIME explanations \cite{vanwijk2023predicting}. Such features, derived from continuous data, could provide a granular view of HRV changes over time, potentially enhancing the predictive capability for clinical deterioration in sepsis patients.

% \textbf{Predictive Accuracy and Clinical Deterioration:} The study's emphasis on the predictive accuracy of HRV features for clinical deterioration is mirrored in the Neural Network model's LIME explanations, which showcase specific HRV features (vlmax, Teo) with high importances. These features might be closely related to the physiological underpinnings of sepsis progression and could serve as viable indicators for early detection and monitoring, aligning with the study's conclusions on the potential of HRV measurements in the emergency department (ED).

Previous research, in conjunction with our LIME explanations, underscore the potential of HRV analysis in monitoring clinical deterioration in sepsis patients. We advocate for the adoption of HRV features in automated sepsis detection; we argue that the equipment and capabilities required are already in place, as HRV features can be extracted from devices such as ECG and passed to machine learning pipelines for patient diagnosis. As a non-invasive, real-time indicator of physiological change, HRV is a prime feature for risk stratification and monitoring tools in clinical settings. 

\section{Conclusion}
This research effectively leverages heart rate variability to enhance sepsis detection. We pinpoint a critical set of HRV features by implementing a hybrid Boruta-bootstrapping feature engineering mechanism. We assemble a suite of advanced machine learning models with differential $\beta$ values, then ensemble high-precision and high-recall models to visualize a 2-D feature space of probabilities. Though ensembling does not boost model performance significantly, we find that multi-layer neural network models result in the highest F1 score, with our Neural Network achieving an F1 Score of 0.805, a precision of 0.851, and a recall of 0.763. Furthermore, we include Local Interpretable Model-agnostic Explanations to elevate this model from a mere predictive tool to a transparent system with clinical applicability. Consequently, this research marks a significant advancement in sepsis risk stratification, offering healthcare providers a tool that identifies at-risk patients and provides insights into physiological indicators, merging early detection with actionable intelligence. 

\section{Additional Information}
The authors have made the code used in this research available on \href{https://github.com/balajsai/HRVbasedSepsisDetection}{GitHub} \cite{balaji2024}.

% \printbibliography

\clearpage

\section*{Appendix}
\begin{table}[htbp]
\caption{Model Performance Metrics, Beta = 1.0}
\begin{center}
\begin{tabular}{|p{1cm}|l|c|c|c|}
\hline
\textbf{Model}&\textbf{Features selected}&\textbf{Fl Score}&\textbf{Precision}&\textbf{Recall}\\\hline
\multirow{5}*{XGBoost} &bootstrap &\textbf{0.745}& 0.760 &0.731\\\cline{2-5}
&boruta\_bootstrap & \textbf{0.742}& 0.726&0.759\\\cline{2-5}
&boruta & \textbf{0.732}& 0.773& 0.694\\\cline{2-5}
&xgboost &\textbf{0.731} &0.809&0.667\\\cline{2-5}
&randomforest & \textbf{0.726}& 0.695 &0.759\\\hline
\multirow{5}{1cm}{Random Forest}&boruta\_bootstrap &\textbf{0.633}& 0.636& 0.630\\\cline{2-5}
&boruta & \textbf{0.631}& 0.571 &0.704\\\cline{2-5}
&bootstrap &\textbf{0.628}&0.567& 0.704\\\cline{2-5}
&xgboost &\textbf{0.628}&0.567& 0.704 \\\cline{2-5}
&randomforest &\textbf{0.619} &0.600 &0.639\\\hline
\end{tabular}
\label{F1}
\end{center}
\end{table}

\begin{table}[htbp]
\caption{Model Performance Metrics, Beta = 0.5}
\begin{center}
\begin{tabular}{|p{1cm}|l|c|c|c|c|}
\hline
\textbf{Model}&\textbf{Features selected}&\textbf{Fl Score}&\textbf{Precision}&\textbf{Recall}\\\hline
\multirow{5}*{XGBoost} &bootstrap &0.597 &\textbf{1.000}& 0.426\\\cline{2-5}
&boruta\_bootstrap &0.629 &\textbf{0.980}& 0.463\\\cline{2-5}
&boruta &0.675 &\textbf{0.934}& 0.528\\\cline{2-5}
&xgboost &0.651 &\textbf{0.931}& 0.500\\\cline{2-5}
&randomforest &0.674 &\textbf{0.836}& 0.565\\\hline
\multirow{5}{1cm}{Random Forest} &xgboost &0.551 &\textbf{0.780}& 0.426\\\cline{2-5}
& bootstrap &0.556 &\textbf{0.770}& 0.435\\\cline{2-5}
&boruta &0.583 &\textbf{0.761}& 0.472\\\cline{2-5}
&boruta\_bootstrap &0.580 &\textbf{0.750} & 0.472\\\cline{2-5}
&randomforest &0.573 &\textbf{0.729}& 0.472\\\hline
\end{tabular}
\label{Precision}
\end{center}
\end{table}

\begin{table}[htbp]
\caption{Model Performance Metrics, Beta = 2.0}
\begin{center}
\begin{tabular}{|p{1cm}|l|c|c|c|}
\hline
\textbf{Model}&\textbf{Features selected}&\textbf{Fl Score}&\textbf{Precision}&\textbf{Recall}\\\hline
\multirow{5}{1cm}{Random Forest} &boruta\_bootstrap &0.505 &0.350 &\textbf{0.907}\\\cline{2-5}
&boruta &0.511 &0.357 &\textbf{0.898}\\\cline{2-5}
&bootstrap &0.501 &0.348&\textbf{0.899}\\\cline{2-5}
&randomforest &0.500 &0.348 &\textbf{0.889} \\\hline
\multirow{5}*{XGBoost}&boruta\_bootstrap &0.689 &0.574 &\textbf{0.861}\\\cline{2-5}
 &bootstrap &0.679 &0.560 &\textbf{0.861}\\\cline{2-5}
&boruta &0.709 &0.622 &\textbf{0.824}\\\cline{2-5}
&randomforest &0.693 &0.597 &\textbf{0.824}\\\cline{2-5}
&xgboost &0.725 &0.659 &\textbf{0.806}\\\hline
Random Forest&xgboost &0.587& 0.474 &\textbf{0.769}\\\hline
\end{tabular}
\label{Recall}
\end{center}
\end{table}

\onecolumn
\begin{table}[htbp]
\caption{Sepsis Feature Selection and Importance }
\begin{center}
\begin{tabular}{|l|l|p{0.45cm}|p{0.45cm}|p{0.45cm}|p{0.45cm}|p{0.45cm}|}\hline
\textbf{Feature Name} & \textbf{Description} & \textbf{XGB} & \textbf{RF} & \textbf{B} & \textbf{Bo} & \textbf{B\_Bo}\\\hline
{Mean.rate} & {heart rate in BPM} & Y & Y & Y & Y & Y      \\\hline
{Coefficient.of.variation} & {ratio of the standard deviation to the mean} & Y & N & Y & Y & Y      \\\hline
{Poincar..SD1} & {Poincaré plot standard deviation perpendicular the line of identity} & Y & Y & Y & Y & Y      \\\hline
{Poincar..SD2} & {Poincaré plot standard deviation along the line of identity} & Y & Y & Y & Y & Y      \\\hline
{LF.HF.ratio.LS} & {Ratio of LF-to-HF power} & Y & Y & Y & Y & Y      \\\hline
{LF.Power.LS} & {power of the low-frequency band} & Y & Y & Y & Y & Y      \\\hline
{HF.Power.LS} & {power of the high-frequency band} & Y & Y & Y & Y & Y      \\\hline
{DFA.Alpha.1} & {Detrended fluctuation analysis, which describes short-term fluctuations} & Y & N & Y & Y & Y      \\\hline
{DFA.Alpha.2} & {Detrended fluctuation analysis, which describes long-term fluctuations} & N & N & Y & Y & Y      \\\hline
{Largest.Lyapunov.exponent} & {measures a non-linear system's sensitive dependence on starting   conditions} & N & N & N & Y & N      \\\hline
{Correlation.dimension} & {Estimate of required model variables}& Y & N & Y & Y & Y      \\\hline
{Power.Law.Slope.LS} & {Power Law (based on frequency) slope $x^2$} & Y & Y & Y & Y & Y      \\\hline
{Power.Law.Y.Intercept.LS} & {Power Law (based on frequency) $y$-intercept $x^2$} & Y & Y & Y & Y & Y      \\\hline
{DFA.AUC} & {Detrended fluctuation analysis: area under the curve} & Y & Y & Y & Y & Y      \\\hline
{Multiscale.Entropy} & {Multiscale entropy - measures the regularity and complexity of a time   series} & Y & N & Y & N & N\\\hline
{VLF.Power.LS} & {Absolute power of the very-low-frequency band} & Y & Y & Y & Y & Y      \\\hline
{Complexity} & {Hjorth parameter complexity} & Y & Y & Y & Y & Y      \\\hline
{eScaleE} & {Embedding scaling exponent} & N & N & N & Y & N      \\\hline
{pR} & {Recurrence quantification analysis: percentage of recurrences} & Y & N & N & Y & N      \\\hline
{pD} & {Recurrence quantification analysis: percentage of determinism} & Y & N & N & N & N      \\\hline
{dlmax} & {Recurrence quantification analysis: maximum diagonal line} & Y & N & N & Y & N      \\\hline
{sedl} & {Recurrence quantification analysis: Shannon entropy of the diagonals} & Y & N & N & Y & N      \\\hline
{pDpR} & {Recurrence quantification analysis: determinism/recurrences} & Y & N & N & Y & N      \\\hline
{pL} & {Recurrence quantification analysis: percentage of laminarity} & N & N & N & N & N      \\\hline
{vlmax} & {Recurrence quantification analysis: maximum vertical line} & Y & N & Y & Y & Y      \\\hline
{sevl} & {Recurrence quantification analysis: Shannon entropy of the vertical   lines} & N & N & N & N & N      \\\hline
{shannEn} & {Shannon entropy, measures uncertainty in a random variable} & Y & Y & Y & Y & Y      \\\hline
{PSeo} & {Plotkin and Swamy energy operator: average energy} & Y & Y & Y & Y & Y      \\\hline
{Teo} & {Teager energy operator: average energy} & Y & Y & Y & Y & Y      \\\hline
{SymDp0\_2} & {Symbolic dynamics: percentage of 0 variations sequences, non-uniform   case} & Y & N & N & Y & N      \\\hline
{SymDp1\_2} & {Symbolic dynamics: percentage of 1 variations sequences, non-uniform case} & Y & N & N & Y & N      \\\hline
{SymDp2\_2} & {Symbolic dynamics: percentage of 2 variations sequences, non-uniform case} & Y & N & Y & Y & Y      \\\hline
{SymDfw\_2} & {Symbolic dynamics: forbidden words, non-uniform case} & Y & N & N & Y & N      \\\hline
{SymDse\_2} & {Symbolic dynamics: Shannon entropy, non-uniform case} & Y & N & N & Y & N      \\\hline
{SymDce\_2} & {Symbolic dynamics: modified conditional entropy, non-uniform case} & Y & N & N & Y & N      \\\hline
{formF} & {Form factor} & Y & N & Y & Y & Y      \\\hline
{gcount} & {Grid transformation feature: grid count} & Y & Y & Y & Y & Y      \\\hline
{sgridAND} & {Grid transformation feature: AND similarity index} & Y & Y & Y & Y & Y      \\\hline
{sgridTAU} & {Grid transformation feature: time delay similarity index} & Y & N & Y & Y & Y      \\\hline
{sgridWGT} & {Grid transformation feature: weighted similarity index} & Y & N & Y & Y & Y      \\\hline
{aFdP} & {Allan factor distance from a Poisson distribution} & Y & Y & Y & Y & Y      \\\hline
{fFdP} & {Fano factor distance from a Poisson distribution} & Y & Y & Y & Y & Y      \\\hline
{IoV} & {Index of variability distance from a Poisson distribution} & Y & Y & Y & Y & Y      \\\hline
{KLPE} & {Kullback-Leibler permutation entropy} & Y & Y & Y & Y & Y      \\\hline
{AsymI} & {Multiscale time irreversibility asymmetry index} & Y & Y & Y & Y & Y      \\\hline
{CSI} & {Poincaré plot cardiac sympathetic index} & Y & N & Y & Y & Y      \\\hline
{CVI} & {Poincaré plot cardiac vagal index} & Y & Y & Y & Y & Y      \\\hline
{ARerr} & {Predictive feature: error from an autoregressive model} & Y & Y & Y & Y & Y      \\\hline
{histSI} & {Similarity index of the distributions} & Y & N & N & Y & N      \\\hline
{MultiFractal\_c1} & {Multifractal spectrum cumulant of the first order} & Y & N & Y & Y & Y      \\\hline
{MultiFractal\_c2} & {Multifractal spectrum cumulant of the second order} & Y & N & Y & Y & Y      \\\hline
{SDLEalpha} & {Scale-dependent Lyapunov exponent slope} & Y & N & Y & Y & Y      \\\hline
{SDLEmean} & {Scale-dependent Lyapunov exponent mean value} & Y & N & Y & Y & Y      \\\hline
{QSE} & {Quadratic sample entropy} & Y & Y & Y & Y & Y      \\\hline
{Hurst.exponent} & {Rate at which autocorrelations decrease as the lag between pairs of   values increases} & Y & Y & Y & Y & Y      \\\hline
{mean} & {Mean value} & Y & N & Y & Y & Y      \\\hline
{median} & {Median value} & Y & N & Y & Y & Y\\\hline
\multicolumn{7}{|p{0.9\textwidth}|}{XGB refers to XGBoost, RF refers to Random Forest, B refers to Boruta, Bo refers to Bootstrap, and B\_Bo refers to Boruta-bootstrap, each representing different feature selection methodologies applied in this study.
}\\\hline
\end{tabular}
\end{center}
\label{Feature_sel}
\end{table}
\twocolumn

\end{document}